\documentclass[lettersize,journal]{IEEEtran}
\usepackage{amsmath,amsfonts}
\usepackage{algorithmic}
\usepackage{algorithm}
\usepackage{array}
\usepackage[caption=false,font=normalsize,labelfont=sf,textfont=sf]{subfig}
\usepackage{textcomp}
\usepackage{stfloats}
\usepackage{url}
\usepackage{verbatim}
\usepackage{graphicx}
\usepackage{cite}
\usepackage{siunitx}
\usepackage{color} 
\usepackage{amsmath}

\DeclareMathOperator{\atantwo}{atan2}

\usepackage{adjustbox}
\usepackage{booktabs}
\usepackage{acronym}
\acrodef{ncc}[NCC]{Normalized Cross Correlation}
\acrodef{imu}[IMU]{Inertial Measurement Unit}
\acrodef{lio}[LIO]{LiDAR-Inertial Odometry}
\acrodef{lo}[LO]{LiDAR Odometry}
\acrodef{lidar}[LiDAR]{Light Detection And Ranging}
\acrodef{icp}[ICP]{Iterative Closest Point}
\acrodef{flu}[FLU]{Front-Left-Up}
\acrodef{pbid}[PBID]{Projection by ID}
\acrodef{pbra}[PBRA]{Projection By Real Angle}
\acrodef{cpbra}[CPBRA]{Corrected Projection By Real Angle}
\acrodef{rmse}[RMSE]{Root Mean Squared Error}
\acrodef{ate}[ATE]{Absolute Trajectory Error}
\acrodef{re}[RE]{Relative Error}

\usepackage{gensymb}

\usepackage{rigidnotation}
\NewRigidNotation{Vel}{v} 
\NewRigidNotation{Acc}{a} 
\NewRigidNotation{Bias}{b} 
\NewRigidNotation{Gravity}{g} 
\NewRigidNotation{State}{x} 
\SetConciseNotation{\BooleanFalse}

\begin{document}

\title{PG-LIO: Photometric-Geometric fusion for Robust LiDAR-Inertial Odometry
}

\author{Nikhil Khedekar and Kostas Alexis

\thanks{This material was supported by the Research Council of Norway under project REDHUS (grant number 317773).}

\thanks{The authors are with the Autonomous Robots Lab, Norwegian University of Science and Technology (NTNU), O. S. Bragstads Plass 2D, 7034, Trondheim, Norway {\tt\small nikhil.v.khedekar@ntnu.no}}
}



\maketitle

\begin{abstract}
\ac{lio} is widely used for accurate state estimation and mapping which is an essential requirement for autonomous robots. Conventional \ac{lio} methods typically rely on formulating constraints from the geometric structure sampled by the LiDAR. Hence, in the lack of geometric structure, these tend to become ill-conditioned (degenerate) and fail. Robustness of \ac{lio} to such conditions is a necessity for its broader deployment. To address this, we propose PG-LIO, a real-time \ac{lio} method that fuses photometric and geometric information sampled by the LiDAR along with inertial constraints from an \ac{imu}. This multi-modal information is integrated into a factor graph optimized over a sliding window for real-time operation. 
We evaluate PG-LIO on multiple datasets that include both geometrically well-conditioned as well as self-similar scenarios. Our method achieves accuracy on par with state-of-the-art \ac{lio} in geometrically well-structured settings while significantly improving accuracy in degenerate cases including against methods that also fuse intensity. Notably, we demonstrate only $\sim\qty{1}{m}$ drift over a $\qty{1}{km}$ manually piloted aerial trajectory through a geometrically self-similar tunnel at an average speed of $7.5m/s$ (max speed $\qty{10.8}{m/s}$). For the benefit of the community, we shall also release our source code \url{https://github.com/ntnu-arl/mimosa}
\end{abstract}


\section{Introduction}\label{sec:intro}
\ac{lidar} sensors have been widely used for odometry estimation due to their accuracy and wide applicability. Due to the typically low rate of azimuth rotation for mechanically spinning \ac{lidar}s, these are typically complemented with an \ac{imu}. \ac{lidar}-Inertial systems have seen wide adoption being used across robotic embodiments \cite{ebadippp} that are able to afford them in terms of size, weight,  power and compute. 

The most common approach to utilizing the geometric \ac{lidar} data is based on the \ac{icp} method~\cite{beslMethodRegistration3D1992}. The essence of the method is to iteratively minimize a distance metric between identified correspondences as a nonlinear optimization problem. While this works well in geometrically well-structured environments, real-world settings can be geometrically self-similar (e.g. tunnels, fields) and cause the problem to become ill-conditioned (degenerate). 

Alleviating this issue has been the focus of many works tackling it from different directions like - information analysis, constrained optimization and multi-modal fusion\cite{khedekarMIMOSAMultiModalSLAM2022,khattakComplementaryMultiModal2020,zhaoSuperOdometryIMUcentric2021,shanLVISAMTightlycoupledLidarVisualInertial2021,wenLIVERTightlyCoupled2024,zheng2024fastlivo2fastdirectlidarinertialvisual,tunaInformedConstrainedAligned2024}. A recent work~\cite{tunaInformedConstrainedAligned2024} classified the issue of degeneracy as a missing data problem, hence, to solve this, we need additional data from some source. While there have been approaches that fuse additional sensors such as visual or thermal cameras\cite{khedekarMIMOSAMultiModalSLAM2022,khattakComplementaryMultiModal2020,shanLVISAMTightlycoupledLidarVisualInertial2021,zheng2024fastlivo2fastdirectlidarinertialvisual} or radars\cite{nissov2024degradationresilientlidarradarinertialodometry, cynoh-2025-icra}, incorporating additional sensors typically involves additional complexity, weight and power, the latter of which are a particular concern for micro aerial vehicles. Instead, maintaining the sensing set of only a LiDAR and an IMU, we utilize the intensity to aid in such scenarios. Intensity is typically already provided by modern LiDARs. However, utilization of the intensity information is nontrivial due to the sensor's
low resolution, discrete sampling, strong rolling shutter effect, projection model, and dependence of the value on various environmental, sensor, and sampling characteristics.


Motivated by the above, this letter presents PG-LIO, a real-time, tightly-coupled \ac{lio} method that fuses photometric and geometric information along with inertial cues in a factor graph. The photometric factor proposes a \ac{ncc}-based cost function that is better suited to the task of direct photometric error on intensity images from \ac{lidar} that violate the brightness constancy assumption. Furthermore, we also contribute a self-calibrating projection model to account for inaccuracy in point projection to create images by a mechanically spinning \ac{lidar}. The geometric factor minimizes a point-to-plane error and the IMU measurements sampled within a single azimuth rotation of the LiDAR added as a preintegration factor. PG-LIO is evaluated on multiple datasets demonstrating comparable performance against state-of-the-art \ac{lio} in geometrically well-conditioned settings and significantly increased accuracy in geometrically self-similar environments against COIN-LIO~\cite{pfreundschuhCOINLIOComplementaryIntensityAugmented2023}, the state-of-the-art in \ac{lio} methods that also utilizes intensity information. We shall also provide our source code for the benefit of the community.

The rest of this paper is organized as follows: Section~\ref{sec:related_work} reviews the related work, Section~\ref{sec:approach} details our approach, Section~\ref{sec:evaluation} presents the evaluation and conclusions are drawn in Section~\ref{sec:conclusion}.

\section{Related Work}\label{sec:related_work}
In this section we review the literature in the domain of intensity usage/aiding specifically for LiDAR-Inertial Odometry and refer the reader to~\cite{lee2023lidarodometrysurveyrecent} for a broader survey of \ac{lio}.

While intensity has been used in LiDAR Odometry\cite{guadagninoFastSparseLiDAR2022,digiammarinoMDSLAMMulticueDirect2022,duRealTimeSimultaneousLocalization2023}, usage in LiDAR-Inertial Odometry is less common. The authors in \cite{liIntensityAugmentedLiDARInertialSLAM2022} focus on solid-state \acp{lidar} and in addition to geometric curvature-based edge and planar points~\cite{Zhang-RSS-14} also extract intensity edge points on planar patches of the pointcloud. They maintain a map consisting of intensity edge points and find point-to-line correspondences within it equivalently to their geometric subsystem. 

Building upon a state-of-the-art geometric \ac{lio}, the works in~\cite{xuFASTLIO2FastDirect2021}, \cite{zhangRILIOReflectivityImage2023} add reflectivity information to the points in their geometric map. This information is queried when they find the nearest neighbors for points and used to add a cost function to minimize the reflectivity difference. Intensity gradients, however, are typically quite local, and the low density of the map ($0.2~\textrm{m}$) is unable to record these. Additionally, they also propose a projection model for the \ac{lidar} to account for the irregular beam spacings. However, this model does not account for the range-dependent error bias of projection. 

COIN-LIO \cite{pfreundschuhCOINLIOComplementaryIntensityAugmented2023} builds upon \cite{zhangRILIOReflectivityImage2023} and instead operates on patches extracted from the intensity image. To be able to function with the range-dependent intensity instead of the preprocessed reflectivity, they first propose a filtering approach to improve the brightness consistency of the image. Additionally, they propose degeneracy-based informative patch selection which allows for fusion of complementary information from the two modalities. As compared to this work, we propose two key differences that allow our method to outperform the state-of-the-art: a) using a projection model that is closer to the actual construction of the \ac{lidar} and b) using a new formulation of a factor that is more tolerant of the noise in the intensity.

\section{Approach}\label{sec:approach}
\subsection{Notation}
Let $W$, $I$, $L$ represent the World-, \ac{imu}- and \ac{lidar}- frames respectively. $(\cdot)_t$ is used to denote the quantity $(\cdot)$ at the time $t$. $W$ coincides with $I$ at initialization, i.e. $W = I_0$. A point $\mathbf{p}$ in frame $A$ is denoted by $\mathbf{p}_A$. The rotation from frame $B$ to $A$ is denoted by $\mathbf{R}_{AB}$. 
The homogeneous transformation $\mathbf{T}_{WI} = \begin{bmatrix}
    \mathbf{R}_{WI} & \mathbf{p}_{WI} \\
    \mathbf{0}_{1 \times 3} & 1
\end{bmatrix}$ transforms a point $\mathbf{p}_I$ in frame $I$ to frame $W$, $\mathbf{p}_W = \mathbf{T}_{WI} \mathbf{p}_I$. The \ac{imu} and \ac{lidar} are assumed to be synchronized and have a fixed extrinsic transformation $\mathbf{T}_{IL}$ that is calibrated apriori. A state $\mathbf{x}_t$ at time $t$ denoted by
\begin{equation}
    \mathbf{x}_t = (\mathbf{T}_{WI_t}, \mathbf{v}_{W,t}, \mathbf{b}_{I,t}) 
\end{equation}
where $\mathbf{v}_{W,t}$ represents the velocity of $I$ in $W$ and $\mathbf{b}_{I,t} = \begin{bmatrix}
    \mathbf{b}_{a,t} & \mathbf{b}_{g,t}
\end{bmatrix}^\top$ represents the biases of the accelerometer and gyroscope. Additionally, we use $\mathbf{g}_W = g \hat{\mathbf{g}}_W$ to denote the gravity vector in $W$. The accelerometer and gyroscope measurements at time $t$ are represented by $\tilde{\mathbf{a}}_t$ and $\tilde{\boldsymbol{\omega}}_t$ respectively.

\subsection{Overall Architecture}
\begin{figure}[ht]
    \centering
    \includegraphics[width=1\linewidth]{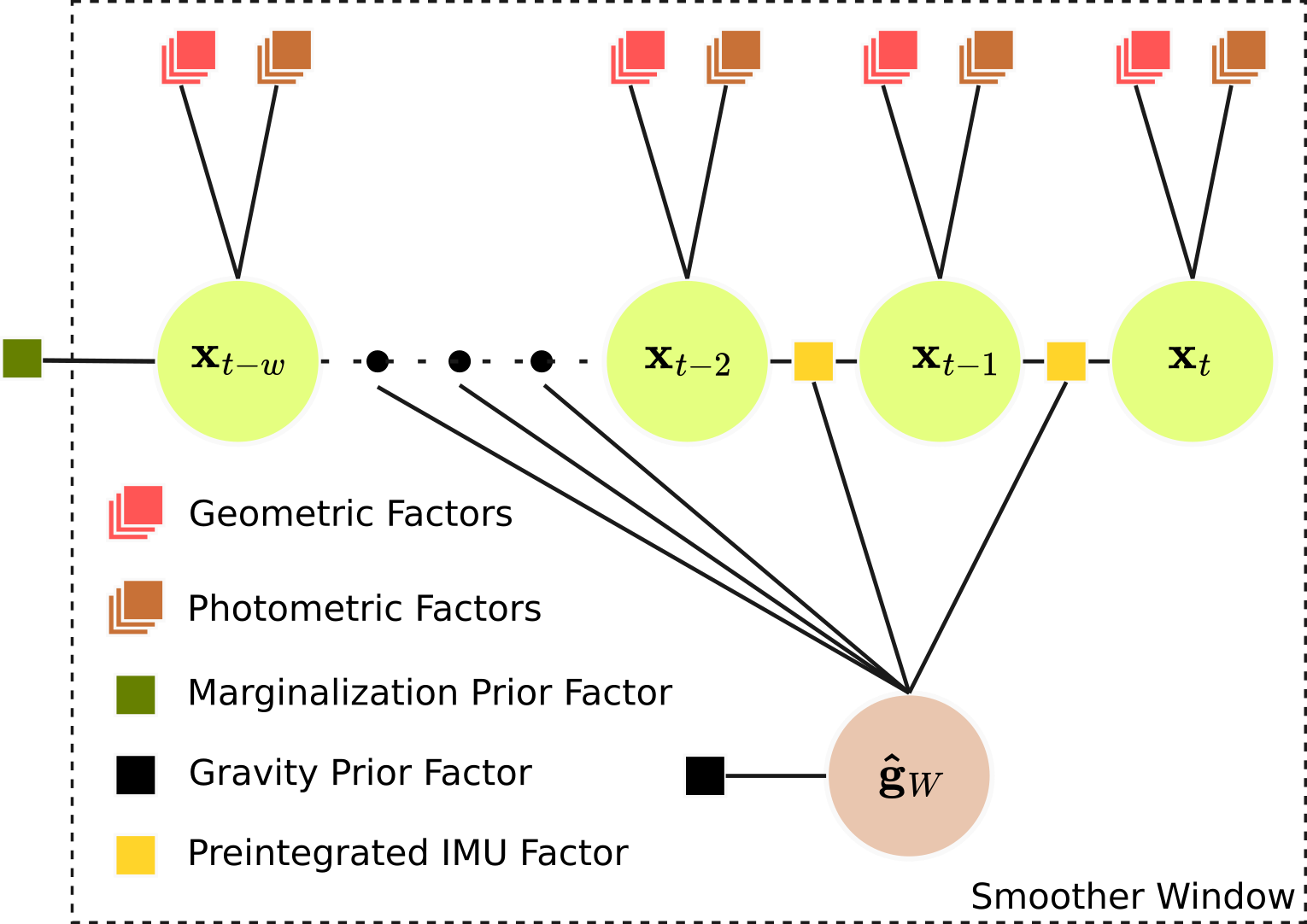}
    \caption{Architecture of the maintained factor graph. Factors are represented by squares while variables are represented with circles.}
    \label{fig:factor_graph}
\end{figure}

Our approach consists of a factor graph that combines factors from Geometric, Photometric and Inertial Subsystems and is smoothed over a sliding window to maintain real-time operation. At time $t$, the variables involved in the factor graph are the robot states timestamped within the sliding window of length $w$, $\mathbf{x}_i \forall i \in [t - w, t]$ as well as the gravity direction $\hat{\mathbf{g}}_W$. The latter is a necessary component due to our assumption of $W = I_0$. The maintained factor graph is visualized in Figure~\ref{fig:factor_graph}.

Overall, an incoming pointcloud is deskewed to correct for motion distortion using \ac{imu} measurements and provided to the Geometric and Photometric subsystems. The factors generated from the individual subsystems are then added to the factor graph and optimized using the incremental fixed lag smoother based on~\cite{kaessISAM2IncrementalSmoothing2012}.


\subsection{Initialization}
We assume that the system is initially static for $N_s$ seconds (typically $N_s =  \qty{0.5}{s}$ and with the \ac{imu} measurements at $\qty{100}{Hz}$ results in $N = \num{50}$ samples). $\mathbf{b}_g$ is initialized as the mean of the gyroscope measurements during this period. Motivated by~\cite{luptonVisualInertialAidedNavigationHighDynamic2012}, we set $W = I_0$ and since the gravity magnitude $g \approx \qty{9.81}{ m/s^2}$ is known to a high degree of accuracy around the world~\cite{balmino2012spherical}, only estimate its direction in $W$. Since $\mathbf{b}_a$ is typically small, we initialize $\hat{\mathbf{g}}_W$ to the direction of the mean of the accelerometer measurements $\boldsymbol{\mu}_a$. $\mathbf{b}_a$ is then initialized as the difference of the mean of the accelerometer measurements and the gravity. In summary,

\begin{align}
    \mathbf{b}_{g, 0} &= \frac{1}{N} \sum_{i} \tilde{\mathbf{\omega}}_i \\
    \boldsymbol{\mu}_a &= \frac{1}{N} \sum_{i} \tilde{\mathbf{a}}_i \\
    \hat{\mathbf{g}}_{W,0} &= \frac{\boldsymbol{\mu}_a}{|\boldsymbol{\mu}_a|} \\
    \mathbf{b}_{a,0} &= \boldsymbol{\mu}_a - g \hat{\mathbf{g}}_{W,0}
\end{align}

Both the geometric and photometric subsystems maintain their maps in $W$ and hence use the first pointcloud unmodified (without deskewing) for initializing these. For the geometric subsystem this is initializing a iVox \cite{baiFasterLIOLightweightTightly2022} map while for the photometric subsystem, this simply implies extracting patch features and storing them in an array.

\subsection{Inertial Subsystem}
Inertial measurements sampled in the duration of the pointcloud are preintegrated to create \ac{imu} preintegration factors. The \ac{imu} factor in~\cite{forsterIMUPreintegrationManifold} constrains not just the relative transformation between the two states but also their absolute states with respect to gravity in $W$ and assumes the gravity to point exactly downward i.e. $\mathbf{g}_W = [0,0,-1]^\top$. Since our $W$ is not necessarily aligned with gravity, the \ac{imu} factor would have a strong corrective gradient that is inconsistent with the photometric and geometric factors. This also becomes problematic when there is very little excitation in rotation in either roll or pitch. Since a highly accurate estimate of the gravity magnitude is typically available~\cite{balmino2012spherical}, we only estimate the direction on $S^2$. The preintegration factor is modified from\cite{forsterIMUPreintegrationManifold} to additionally treat the gravity direction as a variable to be estimated.

\subsection{Geometric Subsystem}
An incoming pointcloud sampled during $[t_s, t_e]$ is first deskewed to $t_e$ by propagating the previous state using the IMU measurements collected during this time and transformed using known extrinsics into $I_e$. The pointcloud is then subsampled for computational efficiency and used to create a point-to-plane geometric factor against a monolithic map~\cite{baiFasterLIOLightweightTightly2022}. If the optimized pose is sufficiently spatially separated in translation or rotation from poses used to update the map, then the new pointcloud is transformed into $W$ and used to update the map. These steps are explained in detail below.

\paragraph{Deskewing}
The \ac{imu} is sampled with a period of $\Delta t$ seconds. As in~\cite{forsterIMUPreintegrationManifold}, we assume that the \ac{imu} measurements  $\tilde{\boldsymbol{\omega}}_t$, $\tilde{\boldsymbol{a}}_t$ at time $t$ represent constant angular velocity and linear acceleration during the period $[t,t+\Delta t]$. Therefore, using $\mathbf{x}_t$, $\tilde{\boldsymbol{\omega}}_t$ and  $\tilde{\boldsymbol{a}}_t$, we can get $\mathbf{x}_{t + \Delta t}$ with

\begin{equation}\label{eq:imu_propagation}
\begin{aligned}
\mathbf{R}_{WI_{t+\Delta t}} &= \mathbf{R}_{WI_t} \operatorname{Exp}\left(\left(\tilde{\boldsymbol{\omega}_t}-\mathbf{b}_{g,t}-\boldsymbol{\eta}_{\mathrm{gd, t}}\right) \Delta t\right) \\
\mathbf{v}_{W, t+\Delta t} &= \mathbf{v}_{W,t}+g\mathbf{\hat{g}}_W \Delta t \\
& +\mathbf{R}_{WI_t}\left(\tilde{\mathbf{a}_t}-\mathbf{b}_{a,t}-\boldsymbol{\eta}_{\mathrm{ad,t}}\right) \Delta t \\
\mathbf{p}_{WI_{t+\Delta t}} &= \mathbf{p}_{WI_t}+\mathbf{v}_{W,t} \Delta t+\frac{1}{2} g\mathbf{\hat{g}}_W \Delta t^2 \\
& +\frac{1}{2} \mathbf{R}_{WI_t}\left(\tilde{\mathbf{a}_t}-\mathbf{b}_{a,t}-\boldsymbol{\eta}_{\mathrm{ad,t}}\right) \Delta t^2
\end{aligned}
\end{equation}
where $\boldsymbol{\eta}_{gd}, \boldsymbol{\eta}_{ad}$ are covariance of the discrete time noises. For more details we refer the reader to~\cite{forsterIMUPreintegrationManifold}.

Equation~\ref{eq:imu_propagation} is used to propagate the state with \ac{imu} measurements during $[t_s, t_e]$. Additionally, Equation~\ref{eq:imu_propagation} is also used to obtain the states at the sampling times of the \ac{lidar}. These states are then used to deskew each point to the \ac{imu} frame at the end of the scan as

\begin{align}
    \mathbf{T}_{L_e L_t} &= \mathbf{T}_{IL}^{-1} \mathbf{T}_{W I_e}^{-1} \mathbf{T}_{W I_t} \mathbf{T}_{I L} \\
    \mathbf{p}_{I_e} &=  \mathbf{T}_{I L} \mathbf{T}_{L_e L_t} \mathbf{p}_{L_t}
\end{align}
Pairs of $(t, \mathbf{T}_{L_e L_t})$ are stored for usage by the Photometric subsystem. Deskewing also provides a prior for the new state ($\mathbf{x}_e$) that will be optimized.

\paragraph{Subsampling}
The number of points sampled by modern \ac{lidar}s is particularly high and contains significant redundant information. For real time operation, we first drop 3 of every 4 points in the pointcloud as in~\cite{xuFASTLIO2FastDirect2021}. To further reduce computation, a voxelgrid downsampling approach is typically used\cite{xuFASTLIO2FastDirect2021}. However, since this replaces sampled points in a voxel with their centroid, it can have the effect of averaging over edges\cite{vizzoKISSICPDefensePointtoPoint2023}. Hence we use 
a subsampling approach\cite{vizzoKISSICPDefensePointtoPoint2023} by adding points to a voxel only if it contains less than $20$ points and the distance between the new point and all of the points in the voxel is greater than $0.1m$.

\paragraph{Geometric Factor}
Each point in the subsampled cloud is transformed using the prior $\mathbf{T}_{WI_e}$ to $W$ to search for a correspondence in the map. A correspondence is deemed valid if the $5$ closest points in the map lie within a certain distance $d_1$, are non collinear (checked by ensuring that the largest eigenvalue is less than 3 times the second largest eigenvalue of the distribution of points) and lie within a threshold $d_2$ of the plane that they represent. Each correspondence contributes a cost $e$ and a Jacobian $\mathbf{J}$ to the optimization. 

\begin{align}
e &= \mathbf{n}_W \cdot (\mathbf{T}_{WI_e} \mathbf{p}_{I_e} - \mathbf{q}_W) \\
\mathbf{J} &= \mathbf{n}_W^T \begin{bmatrix}-\mathbf{R}_{WI_e}\mathbf{p}_{I_e}^\times & \mathbf{R}_{WI_e}\end{bmatrix} \\
&= \begin{bmatrix}(\mathbf{p}_{I_e} \times \mathbf{n}_{I_e})^T & \mathbf{n}_{I_e}^T\end{bmatrix}
\end{align}
where $\mathbf{p}_{I_e}$ is a point in the subsampled pointcloud and $\mathbf{n}_W, \mathbf{q}_W$ are the normal and point lying on the corresponding plane in the map. For computational efficiency, each of these terms is whitened, weighted using a huber kernel~\cite{huber1964robustcostfunction}, and combined to introduce a single dense factor into the factor graph. 

Similar to~\cite{pfreundschuhCOINLIOComplementaryIntensityAugmented2023}, we use the projected localizability contributions in the directions of the eigenvectors of the bottom right (translation) corner of $\mathbf{J}^\top \mathbf{J}$ to identify the directions where the sampling of photometric features is required. 

\paragraph{Map management}
This work considers a monolithic map~\cite{baiFasterLIOLightweightTightly2022} that gets updated at spatial intervals. Since the \ac{lidar} involved has a $360\degree$ horizontal field of view, the map is updated when 1) the distance between the current pose and any pose used to update the map is larger than a threshold ($2m$) or 2) if there is a significant roll or pitch variation ($30\degree$) with the closest pose.

\subsection{Photometric Subsystem}
The deskewed and original pointclouds are used to create an intensity image. After preprocessing steps to remove line artifacts and improve brightness consistency, features from the map are projected into the current image to generate \ac{ncc} based patch factors. The features that are tracked remain, while those that lose tracking are deleted. New features are initialized depending on degenerate directions identified by the Geometric Subsystem. 

\subsubsection{Image Generation and Preprocessing}
Since the incoming pointcloud is organized, the intensity information of each point is trivially formed into an image. This approach is called \ac{pbid}\cite{zhangRILIOReflectivityImage2023}. We use the filtering proposed by~\cite{pfreundschuhCOINLIOComplementaryIntensityAugmented2023} to remove line artifacts and regularize the brightness in the image using a brightness map\cite{pfreundschuhCOINLIOComplementaryIntensityAugmented2023}. Specifically, the line signal is isolated using a highpass filter vertically followed by a lowpass filter horizontally and then subtracted from the original image. The brightness values are averaged over a large window and used to compute a filtered image with consistent exposure. The image is then smoothed with a $3\times 3$ Gaussian kernel for noise reduction.

\subsubsection{Projection model}
Each point in the feature is projected into the image using the projection function $\Pi_n(\cdot)$. The projection model used in~\cite{pfreundschuhCOINLIOComplementaryIntensityAugmented2023} to project a $3\textrm{D}$ point $\mathbf{p}_{L_t} = [x,y,z]^\top$ to $2\textrm{D}$ image coordinates $[u, v]^\top$ is given by

\begin{equation}\label{eq:pinhole_projection}
\begin{bmatrix}
	u \\
	v
\end{bmatrix} = \Pi(\mathbf{p}_{L_t}) = 
\begin{bmatrix}
	f_x \atantwo(y, x) + c_x\\
	f_y \arcsin(\frac{z}{R}) + c_y
\end{bmatrix}
\end{equation}
where $f_x = -w/2\pi, f_y = -h/\theta_{fov}, c_x = w/2, c_y = h/2, R = \sqrt{L^2 + z^2}, L = \sqrt{x^2 + y^2} - n$, $w$ and $h$ represents the width and height of the image respectively, $\theta_{fov}$ is the vertical field of view and $n$ is the distance from the origin of the \ac{lidar} to the projection center. This model misses an important aspect of the construction of the \ac{lidar}. Particularly, the laser head located at the projection center has an azimuth offset $\theta_a$ with the ray from the origin to the laser head. For a measured range $R$ by a laser head at an encoder angle $\theta_e$, and elevation angle $\phi$, the cartesian point is obtained as

\begin{align}\label{eq:proj}
x &= R \cos(\theta_{e} + \theta_{a}) \cos(\phi) + n \cos(\theta_{e}) \\
y &= R \sin(\theta_{e} + \theta_{a}) \cos(\phi) + n \sin(\theta_{e}) \\
z &= R \sin(\phi)
\end{align}

\begin{figure}
    \centering
    \includegraphics[width=1\linewidth]{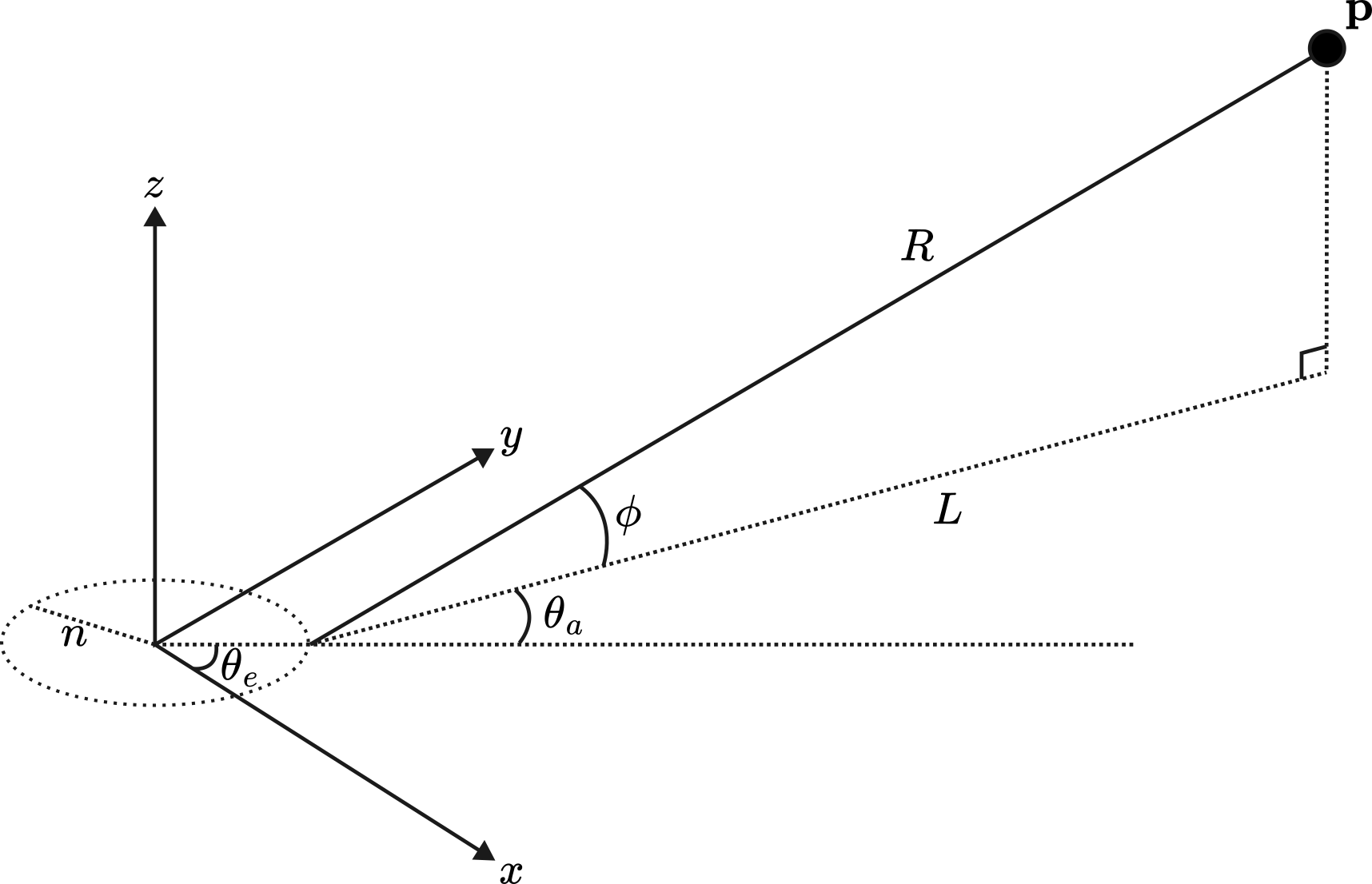}
    \caption{\ac{lidar} Projection model used in the Photometric Subsystem. The model explicitly accounts for the $\theta_a$ offset in the direction of the ray from the origin to the projection center and the ray from the projection center to the sampled point $\mathbf{p}$.}
    \label{fig:projection_model}
\end{figure}

\begin{figure}
    \centering
    \includegraphics[width=1\linewidth]{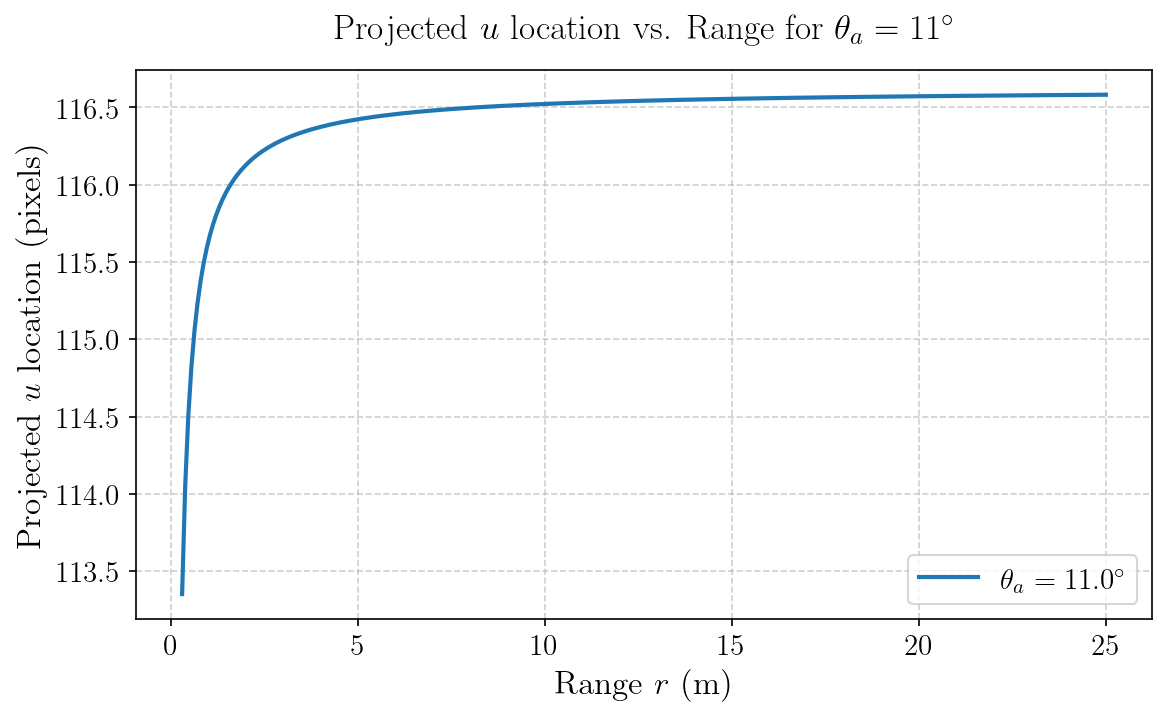}
    \caption{Variation of projected pixel location $u$ with the sampled range $r$ for the projection model in Equation~\ref{eq:pinhole_projection}.}
    \label{fig:projection_location}
\end{figure}
The value of $\theta_a$ can be different for each laser head and depends upon the particular sensor, e.g. in case of Ouster sensors $\theta_a$ can take values in $[-11\degree,+11\degree]$~\cite{ousterSensorData}. This is visualized in Figure~\ref{fig:projection_model}. Therefore, when projecting the point into the image with the model in Equation~\ref{eq:pinhole_projection}, the value of $u$ can vary by several pixels as the measured range varies from $0.3m$ to $20m$. This variation is visualized in Figure~\ref{fig:projection_location} for $\theta_a = 11\degree, n = 0.02767, \theta_e = 30\degree, w = 1024$. While the \ac{cpbra} approach proposed in~\cite{zhangRILIOReflectivityImage2023} models this as a constant bias offset separately for each ring, as can be seen from Figure~\ref{fig:projection_location}, this offset varies with the range of the point. We instead compute the bias offset for each point in the incoming pointcloud to generate a lookup table of bias values to correct the projected locations to be exactly at the \ac{pbid} locations. Different from \ac{cpbra} this accounts for the sampled range. The final projection function is then given by

\begin{equation}
\begin{bmatrix}
	u \\
	v
\end{bmatrix} = \Pi_n(\mathbf{p}_{L_t}) =  \Pi(\mathbf{p}_{L_t}) + \begin{bmatrix}
    b_u \\
    0
\end{bmatrix}
\end{equation}
where $b_u$ is the bias offset obtained from the generated lookup table. In case of invalid points, there will be missing values in the lookup table. This is handled by interpolating between the first valid values in each row. $v$ is further refined to subpixel accuracy by interpolation of the elevation angles of the laser heads\cite{pfreundschuhCOINLIOComplementaryIntensityAugmented2023}.

\subsubsection{Brightness consistency}
The intensity values are a function of environmental and \ac{lidar} parameters, range, surface reflectivity and the angle of incidence~\cite{hewittIntensityaugmentedSLAMLiDAR2015}. Without painstaking sensor-specific calibration, getting intensity images that maintain brightness consistency is particularly difficult\cite{viswanathReflectivityAllYou2024}. While the filtering proposed in~\cite{pfreundschuhCOINLIOComplementaryIntensityAugmented2023} is able to provide significantly greater consistency than the raw intensity image, we find that the consistency is not enough to assume brightness constancy, i.e. two consecutively sampled points cannot be assumed to have the same resulting filtered intensity value.

\ac{ncc} has been used~\cite{woodfordLargeScalePhotometric} in situations where there is an unknown affine change in brightness and hence we use it to formulate our factor. Specifically, our factor is an implementation of the \ac{ncc}~\cite{szeliskiComputerVisionAlgorithms2022} (Eq. 9.11) based on~\cite{woodfordLeastSquaresNormalized2022}. We briefly recap the relevant details due to lack of space and refer the interested reader to~\cite{woodfordLeastSquaresNormalized2022} for a detailed study.

\subsubsection{Recap of NCC}

Given an image $\mathbf{I}$ and a vector of the intensities $\overline{\mathbf{I}} = \mathbf{I}(\mathbf{X}) \in \mathbb{R}^M$ at sampled locations $\mathbf{X} \in \mathbb{R}^{M \times 2}$, a normalization function $\Psi : \mathbb{R}^M \to \mathbb{R}^M$ is defined as

\begin{align}
\Psi(\overline{\mathbf{I}}) &= \frac{\overline{\mathbf{I}}-\mu_{\overline{\mathbf{I}}}}{\sigma_{\overline{\mathbf{I}}}} \\
\mu_{\overline{\mathbf{I}}} &= \frac{\mathbf{1}^{\top} \overline{\mathbf{I}}}{\mathrm{M}} \\
\sigma_{\overline{\mathbf{I}}} &= \left\|\overline{\mathbf{I}}-\mu_{\overline{\mathbf{I}}}\right\|
\end{align}
which has the useful property of transforming the vector such that it is now zero mean and has unit variance. \Ac{ncc} is then defined as a measure of similarity of two vectors $\overline{\mathbf{S}}$ and $\overline{\mathbf{T}}$ as 

\begin{equation}
    E_{NCC}(\overline{\mathbf{S}},\overline{\mathbf{T}}) = \Psi(\overline{\mathbf{S}})^\top \Psi(\overline{\mathbf{T}})
\end{equation}
which is bounded in the range $[-1, 1]$. It is intuitive that similar vectors will have $E_{NCC}>0$. To be used in least squares, an error function can be defined as 

\begin{align}
    E_{ZNSSD} &= || \Psi(\overline{\mathbf{S}}) - \Psi(\overline{\mathbf{T}}) ||^2 \\
    &= 2 - 2E_{NCC}(\overline{\mathbf{S}},\overline{\mathbf{T}})
\end{align}
which will then be bounded on $[0, 4]$ and values over $2$ being obvious outliers. $\Psi()$ is differentiable as presented in~\cite{woodfordLeastSquaresNormalized2022} with the jacobian

\begin{align}
\mathbf{J}_{\Psi}&= \frac{\partial}{\partial \overline{\mathbf{I}}} \Psi(\overline{\mathbf{I}}) \\
&= \underbrace{\frac{\mathbf{I}-\Psi(\overline{\mathbf{I}}) \Psi(\overline{\mathbf{I}})^{\top}}{\sigma_{\overline{\mathbf{I}}}}}_{\text {Variance norm. }} \underbrace{\left(\mathbf{I}-\frac{\mathbf{1 1^\top}}{\mathrm{M}}\right)}_{\text {Zero mean }}
\end{align}

\subsubsection{Map management}
Features in the map exist as patches of points. Candidate patches of high gradient are first extracted from the filtered intensity image. These are then scored and selected on the basis of their localizabilty contributions to the directions detected by the Geometric Subsystem as in~\cite{pfreundschuhCOINLIOComplementaryIntensityAugmented2023}. Once a patch is marked as an outlier in the photometric factor, it is removed from the map.

\subsubsection{Photometric Factor}
All patches in the map are projected into the current image to create the \ac{ncc} based error 

\begin{equation}
    e = \Psi(\mathbf{I}_{j}(\Pi_n(\Gamma(\mathbf{P}(\mathbf{X}))))) - \Psi(\mathbf{I}_k(\mathbf{X}))
\end{equation}
where $\mathbf{I}_k$ is the original image that the feature was first extracted in, $\mathbf{P}$ denotes the corresponding $3D$ points $\mathbf{p}_W$, $\Gamma(\cdot)$ is the relevant homogenous transformation per point, $\Pi(\cdot)$ is the $3D$ to $2D$ projection function and $\mathbf{I}_j$ is the current intensity image. The jacobian is derived through the chain rule as 

\begin{align}
\mathbf{J} &= \frac{\partial e}{\partial \mathbf{T}_{WI_e}} \\
&= \mathbf{J}_\Psi \mathbf{J}_I \mathbf{J}_{\Pi_n} \mathbf{J}_\Gamma \\
\mathbf{J}_{\Pi_n} &= \frac{\partial\Pi_n(\mathbf{p}_{L_t})}{\partial \mathbf{p}_{L_t}} = \begin{bmatrix}
\frac{-f_x y}{x^2 + y^2} & \frac{f_x x}{x^2 + y^2} & 0  \\
\frac{-f_y zx}{R^2 (L+n)} & \frac{-f_y zy}{R^2 (L+n)} & \frac{f_y L}{R^2} 
\end{bmatrix} \\
\mathbf{J}_\Gamma &= \frac{\partial (\mathbf{T}_{L_e L_t}^{-1} \mathbf{T}_{IL}^{-1} \mathbf{T}_{WI_e}^{-1} \mathbf{p}_W)}{\partial \mathbf{T}_{WI_e}} \\
&= \mathbf{R}_{L_t I_e} \begin{bmatrix}(\mathbf{T}_{WI_e}^{-1}\mathbf{p}_W)^\times & -\mathbf{I}_{3 \times 3}\end{bmatrix}
\end{align}

$\mathbf{J}_I$ is simply the image gradient at the sampled locations. Up to the last step of sampling, each of the points is treated individually - only in the last \ac{ncc} step do they all become a patch. The undistortion map proposed in~\cite{pfreundschuhCOINLIOComplementaryIntensityAugmented2023} is used to obtain the correct $\mathbf{T}_{L_e L_t}$ for each $\mathbf{p}_W$. Outliers are rejected by tracking depth differences in the projections and a minimum threshold on $E_{NCC}$. The cost terms for each individual patch are combined to create a single dense photometric factor to be added into the factor graph.

\section{Evaluation}\label{sec:evaluation}
We evaluate our approach on a set of relevant datasets demonstrating not only comparability to state-of-the-art methods in geometrically well-conditioned scenarios but critically also robustness with improved accuracy in geometrically degenerate environments. The values in Table~\ref{tab:res_ncd} and Table~\ref{tab:res_enwide} for ~\cite{vizzoKISSICPDefensePointtoPoint2023,digiammarinoMDSLAMMulticueDirect2022,shanLIOSAMTightlycoupledLidar2020,xuFASTLIO2FastDirect2021,zhangRILIOReflectivityImage2023,pfreundschuhCOINLIOComplementaryIntensityAugmented2023} are as reported by~\cite{pfreundschuhCOINLIOComplementaryIntensityAugmented2023} and hence, following their convention, we also report the \ac{rmse} \ac{ate} in meters and \ac{re} in percent for a delta of $10m$ as evaluated by the open-source $evo$ package\cite{grupp2017evo}.

\subsection{Newer College Dataset}
\begin{figure*}
    \centering
    \includegraphics[width=\linewidth]{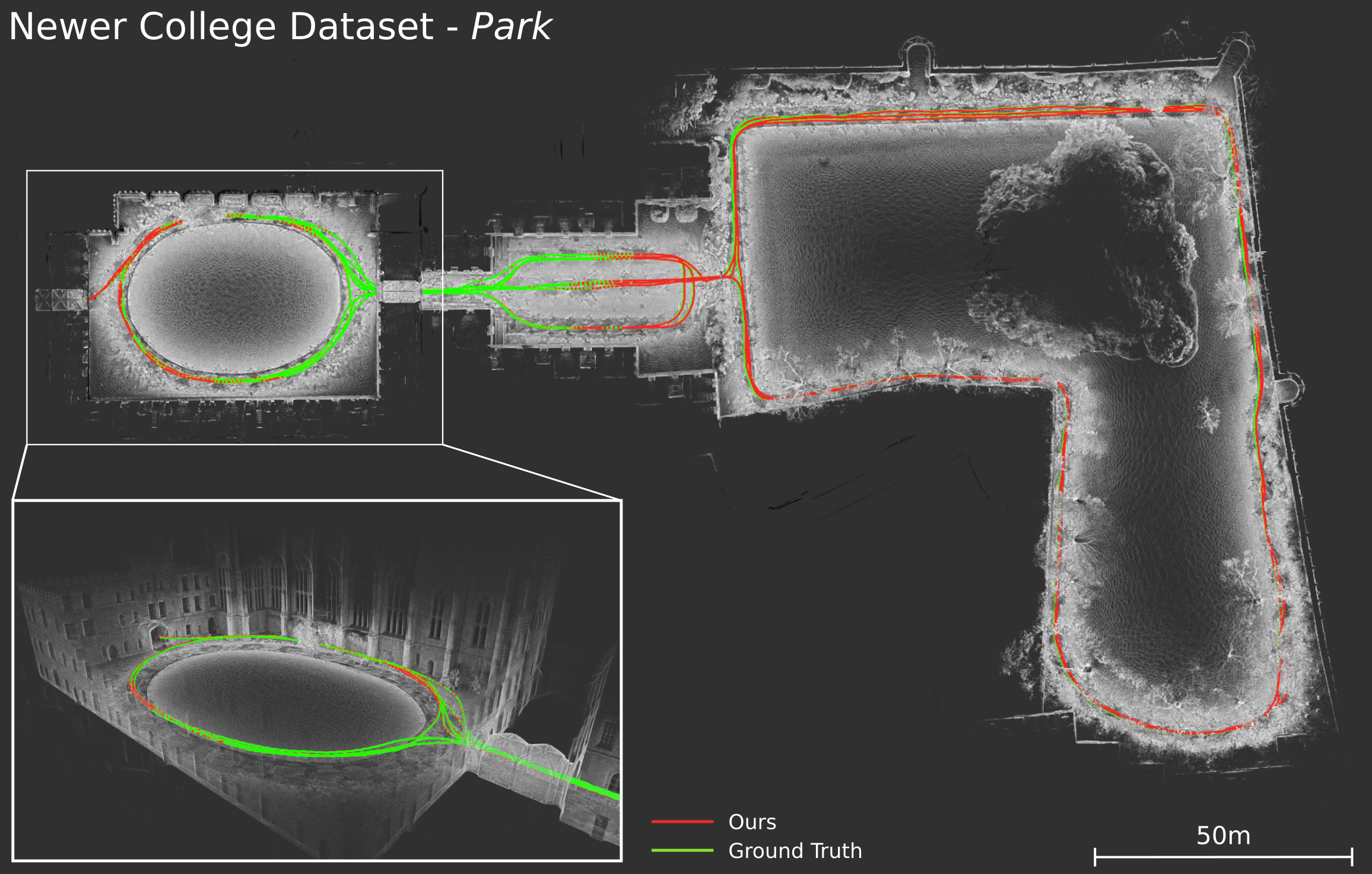}
    \caption{Top-Down Orthographic view of the accumulated registered pointcloud for the $Park$ sequence of the Newer College Dataset. Inset: Close-up isometric view of the initial courtyard showcasing the accuracy of the reconstruction.}
    \label{fig:result_nc}
\end{figure*}

To show the general functioning of our method, we first evaluate our method on some of the sequences in the Newer College Dataset~\cite{zhangMultiCameraLiDARInertial2021} and report the results in Table~\ref{tab:res_ncd}. The dataset consists of a handheld Ouster OS0-128 \ac{lidar} being carried around different sections of a college campus. Our approach performs comparably to state-of-the-art \ac{lio} methods in all sequences. The accumulated registered pointcloud as well as the estimated and ground truth trajectory for the $Park$ sequence are visualized in Figure~\ref{fig:result_nc}.

\begin{table}
\caption{Newer College Dataset \\
Absolute Trajectory Error (\ac{rmse}) (\textit{m}) / Relative Error (\textit{\%})}
\begin{adjustbox}{max width=\linewidth}
\begin{tabular}{ccccc}
\toprule
Method &  Quad-Hard & Cloister & Stairs & Park \\
Length \textit{(m)} & $234.81$ & $428.79$ & $57.04$ & $2396.20$ \\
\midrule
  KISS-ICP~\cite{vizzoKISSICPDefensePointtoPoint2023} & $0.324$ / $1.88$ & $0.297$ / $2.07$  &  $\times$ / $32.48$ & $2.871$ / $1.06$\\
  MD-SLAM~\cite{digiammarinoMDSLAMMulticueDirect2022} & $19.639$ / $12.36$ & $0.360$ / $2.73$ & $0.340$ / $6.21$ & $96.797$ / $23.03$ \\
  Du and Beltrame~\cite{duRealTimeSimultaneousLocalization2023} & $18.506$ / $16.432$ & $59.544$ / $19.274$ & $\times$ / $26.121$ & $\times$ / $42.717$ \\
  LIO-SAM~\cite{shanLIOSAMTightlycoupledLidar2020} & $0.299$ / $2.380$ & $0.145$ / $1.032$ & $\times$ / $5122.320$ & $1.566$ / $2.064$ \\
  FAST-LIO2~\cite{xuFASTLIO2FastDirect2021} & $0.049$ / $\mathbf{0.26}$ & $\mathbf{0.078}$ / $\mathbf{0.23}$ & $\times$  / $3497.22$ & $0.310$ / $0.59$\\
  RI-LIO~\cite{zhangRILIOReflectivityImage2023} & $0.237$ / $1.04$ & $0.285$ / $1.33$ &  $\times$ / $16877.28$ & $89.289$ / $5.00$  \\
  COIN-LIO~\cite{pfreundschuhCOINLIOComplementaryIntensityAugmented2023} & $\mathbf{0.046}$ / $0.29$ & $\mathbf{0.078}$ / $0.28$& $0.102$ / $\mathbf{0.74}$ & $\mathbf{0.287}$ / $\mathbf{0.54}$ \\
  \textbf{Ours} & $0.054$ / $0.35$ & $0.097$ / $0.55$ & $\textbf{0.077}$ / $0.78$ & $0.331$ / $0.61$  \\ 

\bottomrule
\end{tabular}
\end{adjustbox}
\label{tab:res_ncd}
\end{table}

\subsection{ENWIDE Dataset}

\begin{figure*}[ht]
    \centering
    \includegraphics[width=\linewidth]{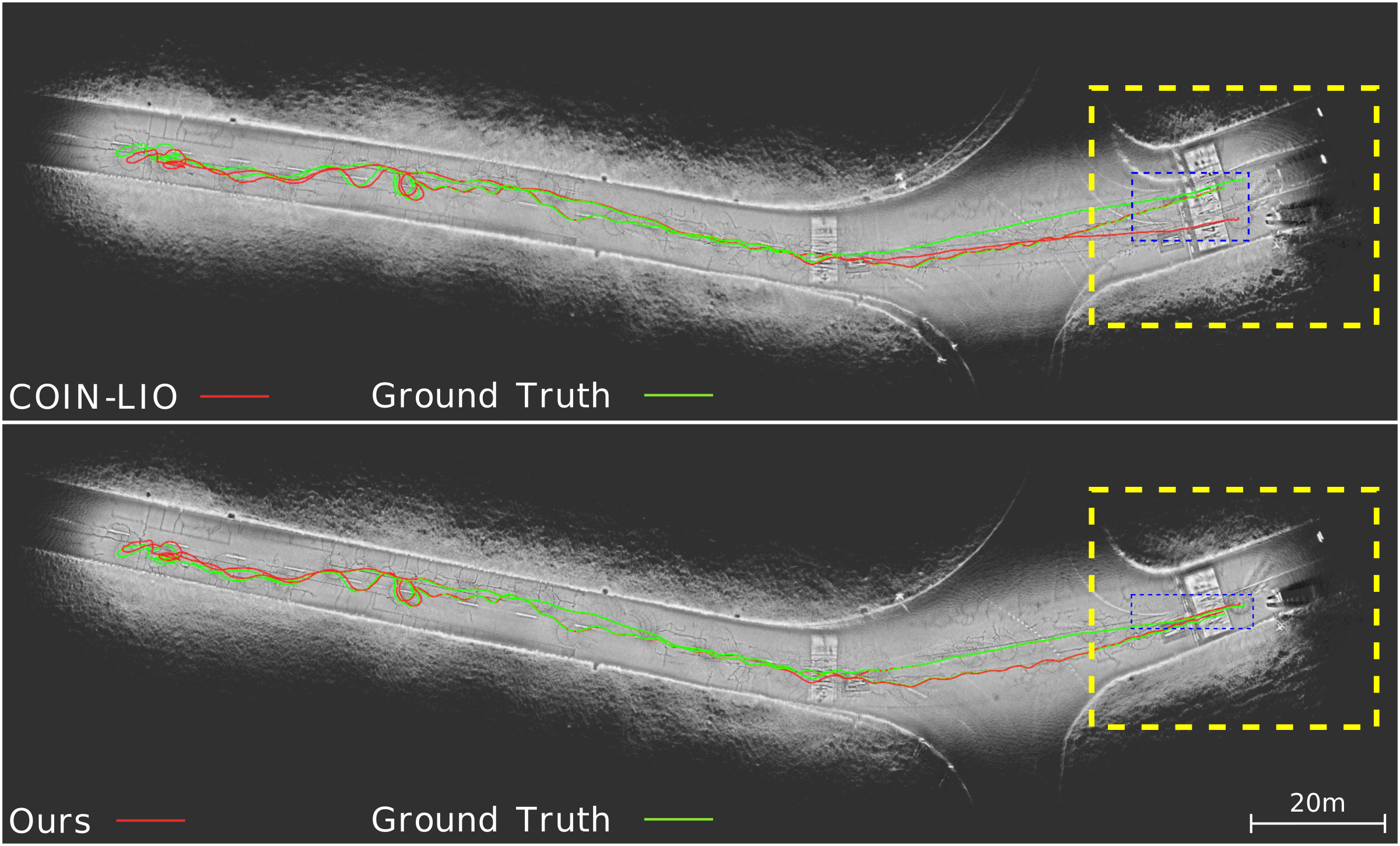}
    \caption{Qualitative comparison demonstrating our improved accuracy over COIN-LIO\cite{pfreundschuhCOINLIOComplementaryIntensityAugmented2023} on the $IntersectionD$ sequence of the ENWIDE dataset. While the section marked in the yellow dashed box shows a lower overall drift achieved by our method, the rest of the map is also more sharp. Additionally, the section in the blue dashed box highlights the overlap of our trajectory against ground truth.}
    \label{fig:chap_pglio/map_comparison}
\end{figure*}

Secondly, we test our method on the ENWIDE Dataset~\cite{pfreundschuhCOINLIOComplementaryIntensityAugmented2023} and present the results in Table~\ref{tab:res_enwide}. This dataset contains several sequences of a handheld OS0-128 \ac{lidar} carried in geometrically degenerate environments with both slow (suffixed with $S$) and dynamic (suffixed with $D$) maneuvers. The ground truth positions are captured by a Leica Total Station for a prism mounted on the \ac{lidar}. 

Of the compared methods, COIN-LIO is the only method that does not completely fail in any sequence. Our method shows a significant improvement over COIN-LIO~\cite{pfreundschuhCOINLIOComplementaryIntensityAugmented2023} in both \ac{ate} and \ac{re} in all sequences except $TunnelS$ and $RunwayD$. In $TunnelS$, this is due to the estimator not converging to the correct gravity and accelerometer bias which causes a significant error on turning at the end of the tunnel. In $RunwayD$, our method suffers a bad estimate of the rotation in a single particularly agile maneuver breaking the map. While a similar break is also seen in \cite{pfreundschuhCOINLIOComplementaryIntensityAugmented2023} at the same time, it happens to deviate again later in the trajectory and returns closer to the initial position resulting in a lower value for the \ac{ate}. Figure~\ref{fig:chap_pglio/map_comparison} shows an indicative result of the improved accuracy by the proposed approach against COIN-LIO on the $IntersectionD$ sequence. 



\begin{table*}[t!]
\caption{ENWIDE Dataset - Absolute Trajectory Error (\ac{rmse}) (\textit{m}) / Relative Error (\textit{\%}) }
\label{tab:res_enwide}
\begin{adjustbox}{max width=\textwidth}
\begin{tabular}{ccccccccccc}
\toprule
Method &  TunnelS & TunnelD & IntersectionS & IntersectionD & RunwayS & RunwayD & FieldS & FieldD & KatzenseeS & KatzenseeD\\ 
Length \textit{(m)} & $251.58$ & $179.71$ & $279.28$ & $388.47$ & $333.57$ & $357.14$ & $232.70$ & $287.91$ & $242.88$ & $177.20$\\
\midrule
  KISS-ICP~\cite{vizzoKISSICPDefensePointtoPoint2023} & $\times$ / $144.41$  & $\times$ / $68.11$ & $\times$ / $65.69$ & $\times$ / $64.84$ & $\times$ / $113.45$ & $\times$ / $124.64$ & $\times$ / $54.84$ & $\times$ / $70.70$ & $\times$ / $66.23$ & $\times$ / $76.80$\\
  MD-SLAM~\cite{digiammarinoMDSLAMMulticueDirect2022} &  $\times$ / $88.16$ &  $\times$ / $80.76$ &  $\times$ / $90.87$ &  $\times$ / $87.89$ &  $\times$ / $97.73$ &  $\times$  / $91.13$&  $\times$ / $96.03$ &  $\times$ / $84.86$ &  $\times$ / $93.92$ &  $\times$ / $91.29$ \\
  Du and Beltrame~\cite{duRealTimeSimultaneousLocalization2023} &  $\times$ / $58.084$ &  $\times$ / $56.086$ &  $\times$ / $60.812$ &  $\times$ / $57.449$ &  $\times$ / $67.906$ &  $\times$  / $63.978$&  $\times$ / $72.548$ &  $\times$ / $69.480$ &  $\times$ / $93.92$ &  $\times$ / $74.207$ \\
  LIO-SAM~\cite{shanLIOSAMTightlycoupledLidar2020} &  $\times$ / $2565.621$ &  $\times$ / $2662.983$ &  $\times$ / $2022.878$ &  $\times$ / $2362.314$ &  $\times$ / $2334.174$ &  $\times$  / $3984.588$&  $\times$ / $2196.344$ &  $\times$ / $1999.968$ &  $5.588$ / $2.673$ &  $\times$ / $1485.377$ \\
  FAST-LIO2~\cite{xuFASTLIO2FastDirect2021} & $\times$ / $316.12$  & $\times$ / $81.31$& $12.473$ / $29.28$  & $23.800$ / $28.11$ & $\times$ / $53.64$ & $\times$ / $59.84$ & $\textbf{0.163}$ / $\textbf{0.57}$ & $9.209$ / $16.08$ & $1.122$ / $4.31$& $1.02$ / $2.38$\\
  RI-LIO~\cite{zhangRILIOReflectivityImage2023} & $\times$ / $70.32$ & $\times$ / $63.02$ & $\times$ / $49.94$& $\times$ / $188.83$& $\times$ / $52.18$& $\times$ / $79.16$ & $1.721$ / $2.44$ & $24.851$ / $25.89$& $\times$ / $49.34$& $\times$ / $154.19$\\
  COIN-LIO~\cite{pfreundschuhCOINLIOComplementaryIntensityAugmented2023} & $0.743$ / $\textbf{1.60}$ & $0.487$ / $1.59$ & $0.466$ / $1.25$ & $1.912$ / $1.69$  & $1.033$ / $1.89$ & $\textbf{2.437}$ / $2.98$  & $0.232$ / $0.85$ & $0.581$ / $1.83$ & $0.412$ / $0.99$ & $0.592$ / $1.61$ \\

\textbf{Ours} & $\textbf{0.635}$ / $2.09$ & $\textbf{0.425}$ / $\textbf{1.45}$ & $\textbf{0.279}$ / $\textbf{0.80}$ & $\textbf{0.473}$ / $\textbf{0.84}$ & $\textbf{0.491}$ / $\textbf{1.00}$ & $3.273$ / $\textbf{1.08}$ & $0.274$ / $0.60$ & $\textbf{0.206}$ / $\textbf{0.88}$ & $\textbf{0.260}$ / $\textbf{0.47}$ & $\textbf{0.355}$ / $\textbf{0.64}$  \\ 

\bottomrule
\end{tabular}
\end{adjustbox}
\end{table*}

\subsection{Geometrically Self-Similar Bicycle Tunnel}
\begin{figure*}[h!]
    \centering
    \includegraphics[width=\linewidth]{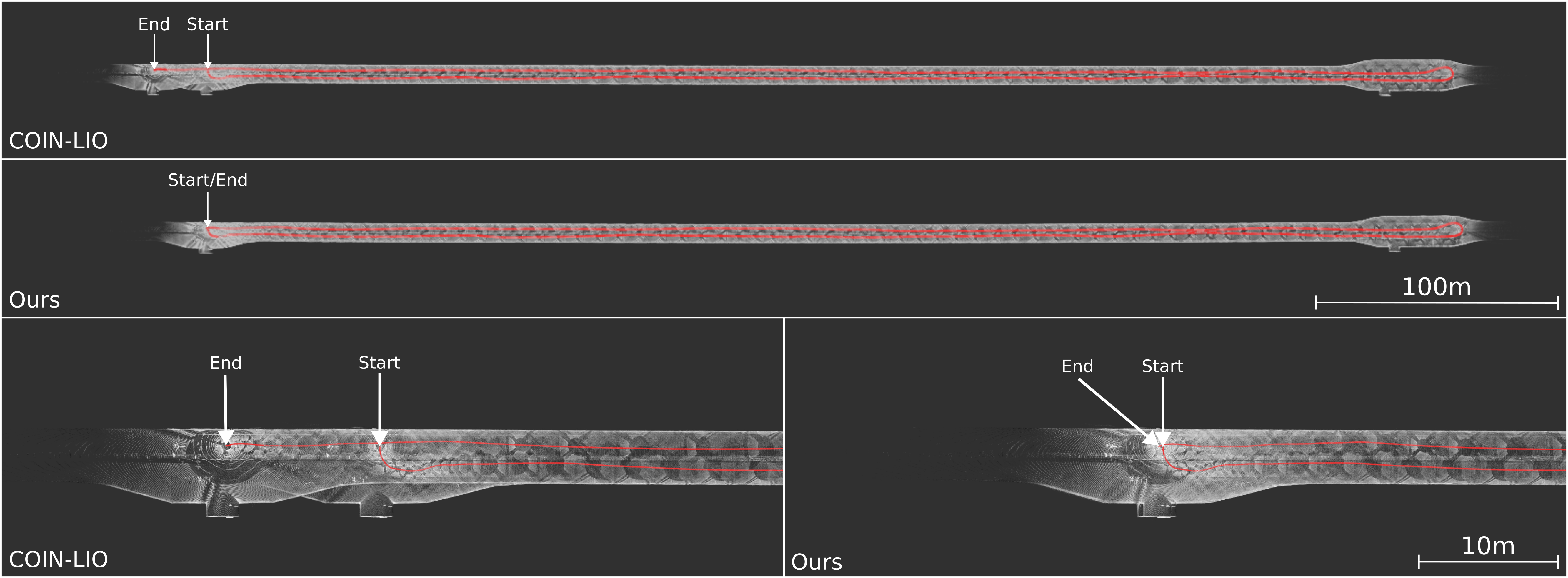}
    \caption{Comparison against COIN-LIO~\cite{pfreundschuhCOINLIOComplementaryIntensityAugmented2023} in a \qty[mode = text]{500}{m} long geometrically self-similar section of a bicycle tunnel collected from an aerial robot manually piloted at an average speed of \qty[mode = text]{7.5}{m/s} and maximum speed of \qty[mode = text]{10.8}{m/s}. The total length of the trajectory is \qty[mode = text]{1047}{m} which is accurately reconstructed by our method.}
    \label{fig:result_fyllingsdal}
\end{figure*}

Finally, we evaluate the proposed method on data collected from an aerial robot that is manually piloted inside a geometrically self-similar section of a bicycle tunnel at high speed. The tunnel contains rest stops every \qty{500}{\meter}, connected by a straight section with a constant semi-circular cross-section of \qty{8}{\meter} diameter.
The tunnel walls contain a mix of plain and decorative mural sections that provide textural cues in the intensity image. The aerial robot, a variant of ~\cite{petrisRMFOwlCollisionTolerantFlying2022}, carries an Ouster OS-0-128 Rev7 \ac{lidar} along with a VectorNav VN100 \ac{imu}. The robot takes off from the first rest stop, flies through the tunnel to the next rest stop and returns to land at the starting location with an overall trajectory length of \qty{1047}{m}. The average speed during the flight is \qty{7.5}{m/s} with the maximum speed achieved is \qty{10.8}{m/s}. Since all of \cite{vizzoKISSICPDefensePointtoPoint2023,digiammarinoMDSLAMMulticueDirect2022,shanLIOSAMTightlycoupledLidar2020,xuFASTLIO2FastDirect2021,zhangRILIOReflectivityImage2023} fail in case of the simpler sequences of the ENWIDE Dataset, we only evaluate our method against COIN-LIO in this experiment. The accumulated registered pointcloud from our method and COIN-LIO is visualized in Figure~\ref{fig:result_fyllingsdal}. While both the methods are able to roughly recreate the length of the tunnel while going from the first to the second rest stop, COIN-LIO loses track of features while on the way back in a visually less featureful region. In comparison, the proposed method returns with a small drift of $\sim\qty{1}{m}$. Though minimal and not easily distinguishable in the reconstructed maps, both methods present a small z-drift. 

\section{Conclusion}\label{sec:conclusion}
This paper presented a real-time, tightly-coupled \ac{lio} method that fuses photometric and geometric information sampled by a \ac{lidar} along with inertial measurements from an \ac{imu}. The method was demonstrated to provide comparable accuracy against the state-of-the-art \ac{lio} in geometrically well-conditioned settings and improved accuracy against the state-of-the-art \ac{lio} that utilizes intensity information in geometrically degenerate environments. The code shall be released open source for the benefit of the research community.


\section*{Acknowledgments}
We would like to thank the Vestland Fylkeskommune for providing access to the Fyllingsdal bicycle tunnel.



\bibliographystyle{IEEEtran}
\bibliography{references/bibliography}


 




\vfill

\end{document}